\documentclass[journal]{IEEEtran} 
\usepackage{amsmath,amsfonts}
\usepackage{algorithmic}
\usepackage{algorithm}
\usepackage{array}
\usepackage[caption=false,font=footnotesize,labelfont=sf,textfont=sf]{subfig}
\usepackage{textcomp}
\usepackage{stfloats}
\usepackage{url}
\usepackage{verbatim}
\usepackage{graphicx}
\usepackage{cite}
\usepackage{multirow}
\usepackage{enumitem}
\usepackage{booktabs}
\usepackage{xspace}
\usepackage{makecell}
\def \alambic {\includegraphics[width=0.013\linewidth]{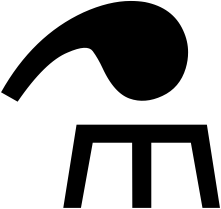}\xspace}

\begin{document}

\title{A Lightweight Target-Driven Network of Stereo Matching for Inland Waterways}

\author{Jing Su, Yiqing Zhou, Yu Zhang, Chao Wang, Yi Wei

\thanks{(Jing Su and Yiqing Zhou contributed equally to this work.) (Corresponding author: Yu Zhang.)}
\thanks{Jing Su, Yiqing Zhou and Yu Zhang are with the College of Artificial Intelligence, Tianjin University of Science and Technology, Tianjin 300457, China (e-mail: sujing@tust.edu.cn, dgar12@mail.tust.edu.cn, zhangyuai@tust.edu.cn).}
\thanks{Chao Wang is with the School of Artificial Intelligence, Anhui University, Hefei 230601, China (e-mail: wangchao8@ahu.edu.cn).}
\thanks{Yi Wei is with the Tianjin Institute of Navigation Instruments, Tianjin 300131, China (e-mail: hityiwei@163.com).}}

\maketitle

\begin{abstract}
Stereo matching for inland waterways is one of the key technologies for the autonomous navigation of Unmanned Surface Vehicles (USVs), which involves dividing the stereo images into reference images and target images for pixel-level matching. However, due to the challenges of the inland waterway environment, such as blurred textures, large spatial scales, and computational resource constraints of the USVs platform, the participation of geometric features from the target image is required for efficient target-driven matching. Based on this target-driven concept, we propose a lightweight target-driven stereo matching neural network, named LTNet. Specifically, a lightweight and efficient 4D cost volume, named the Geometry Target Volume (GTV), is designed to fully utilize the geometric information of target features by employing the shifted target features as the filtered feature volume. Subsequently, to address the substantial texture interference and object occlusions present in the waterway environment, a Left-Right Consistency Refinement (LRR) module is proposed. The \text{LRR} utilizes the pixel-level differences in left and right disparities to introduce soft constraints, thereby enhancing the accuracy of predictions during the intermediate stages of the network. Moreover, knowledge distillation is utilized to enhance the generalization capability of lightweight models on the USVInland dataset. Furthermore, a new large-scale benchmark, named Spring, is utilized to validate the applicability of LTNet across various scenarios. In experiments on the aforementioned two datasets, LTNet achieves competitive results, with only 3.7M parameters. The code is available at \url{https://github.com/Open-YiQingZhou/LTNet}.
\end{abstract}


\section{Introduction}

\IEEEPARstart{I}{nland} waterways, as an essential part of the hydrological cycle, are closely linked to human life and formed the foundation of modern urban and community development. However, the narrow waterways and diverse water surface environments pose numerous challenges for management and protection. To reduce development and operational costs and enhance personnel safety, Unmanned Surface Vehicles (USVs) equipped with stereo vision systems have been developed to perform tasks in various riverine environments \cite{barrera2021trends, cheng2021we}. Similar to the stereo vision systems used in autonomous driving road vehicles and indoor robots, the stereo vision system for USVs identifies dense matching correspondences from rectified stereo image pairs and estimates the disparity map. Subsequently, depth information can be easily derived through triangulation for 3D environmental reconstruction, which is utilized in the automatic navigation of USVs.

\begin{figure}[tbp]
\centering
\def \alambicm {\includegraphics[width=0.025\linewidth]{figs/alembic-crop.pdf}\xspace}
\subfloat{\includegraphics[width=.9\linewidth]{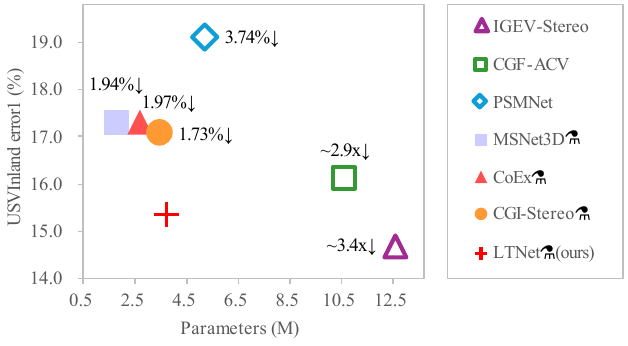}}
\caption{1-pixel error rate and Parameters comparison between our LTNet and other stereo networks on the USVInland dataset. The symbol $\alambicm$ refers to model trained using distillation. It can be seen that the LTNet can achieve high predictive accuracy while maintaining a small model size.}
\label{fig:comparison}
\end{figure}

Stereo matching is the cornerstone of 3D scene understanding. Compared to inland waterways, the research on stereoscopic perception for road vehicles and indoor robots has a long history and rich achievements. The most advanced stereo matching methods for roads and indoor environments currently utilize CNNs to automatically extract features from input left and right images for designing end-to-end networks \cite{kendall2017end, chang2018pyramid, chen2023unambiguous, wu2024towards}. It constructs a 4D cost volume containing abundant matching information by pre-defined disparity shifts, then repeatedly applies extensive 3D convolutions for aggregation. Finally, the soft argmin operation is employed to regress and generate the disparity map. Some studies draw on the successful experience of recurrent neural networks, construct cost volumes based on full correlation, and use ConvGRU to iteratively refine the coarse disparity estimation, such as Raft-Stereo \cite{lipson2021raft} and IGEV-Stereo \cite{xu2023iterative}. Furthermore, some methods adopt patch segmentation techniques to represent images as dense sequences, creating cost volumes through cross-attention mechanisms between stereo pairs, thereby avoiding the maximum disparity limitation of manually crafted cost volumes. These approaches extensively utilize self-attention mechanisms for modeling global information, which enhances the model's ability to match fine details, such as STTR \cite{li2021revisiting}, GMStereo \cite{xu2023unifying}, and CroCo-Stereo \cite{weinzaepfel2023croco}. However, the excessive parameterization of large-scale convolutional methods and attention networks, coupled with the limited hardware resources of USVs in inland waterways, leads to substantial computational and storage demands that often exceed the platform's capacity, posing significant challenges for practical deployment.

On the other hand, in road and indoor scenes, the scenery typically exhibits regularity and predictability, such as traffic markings, vehicles, and indoor man-made objects. This man-made structured environment is characterized by objects and surfaces that adhere to certain geometric rules, which provide rich, exploitable geometric and textural features for stereo matching algorithms. In contrast, unstructured environments, like inland waterways, lack this regularity, and their natural textural features may be more chaotic and blurred, including the irregular edges of river channels, the fluctuation of water surfaces, and the plants near the river. This ambiguity in texture poses a challenge for general lightweight methods to effectively encode costs in environments lacking artificial objects. In terms of spatial scale, according to our statistics, the road KITTI2015 dataset \cite{menze2015object} exhibits a maximum disparity of the ground truth at 229 pixels, whereas the inland waterway USVInland dataset \cite{cheng2021we} shows a significantly smaller maximum disparity of only 50 pixels. This indicates that inland waterways typically have a smaller disparity and larger depth of field, which can make key textural information at greater distances significantly weakened, leading to the blurring of textural features and thus increasing the difficulty of matching.

\begin{figure}[tbp]
\def \alambicm {\includegraphics[width=0.025\linewidth]{figs/alembic-crop.pdf}\xspace}
\subfloat{\includegraphics[width=\linewidth]{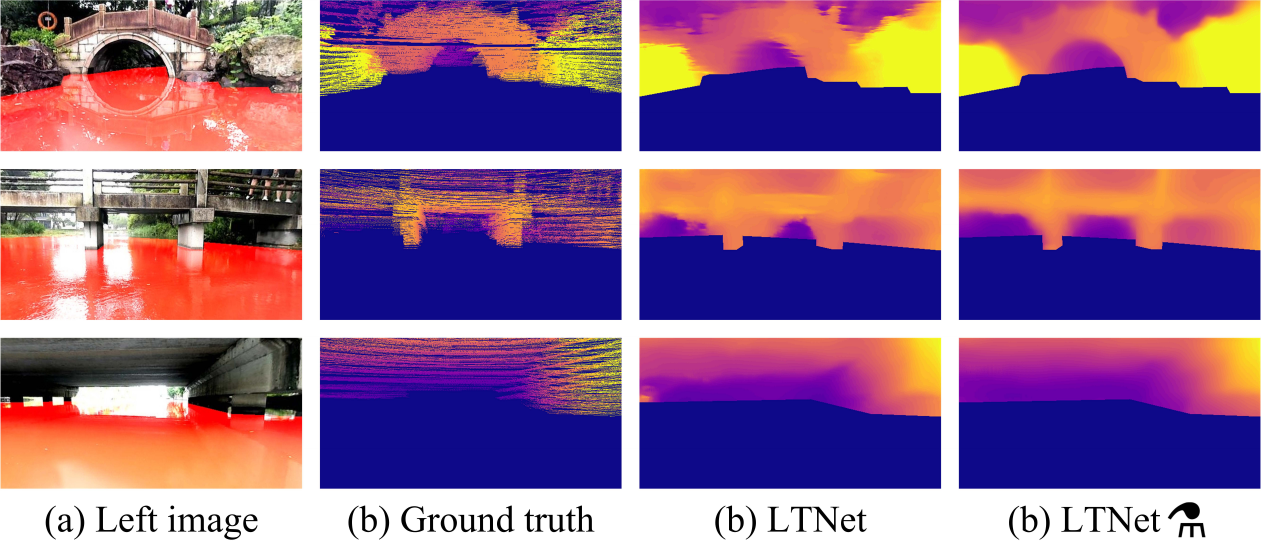}}
\caption{Rectified stereo matching results for 3D perception of inland waterway scenes using our LTNet and the corresponding results obtained through the method of knowledge distillation. The symbol $\alambicm$ refers to model trained using distillation. Please note that our research does not focus on the depth of the water surface. In our demonstrations, water areas have been removed using red and navy blue.}
\label{fig:describe}
\end{figure}

To alleviate the aforementioned challenges, this paper proposes a lightweight target-driven network named LTNet. LTNet incorporates two distinctive designs, driven by the concept of target-driven image matching. One of them is a lightweight 4D cost volume, named Geometric Target Volume (GTV), which constructs a manual cost directly using geometric features extracted from the target image, addressing the issue that cluttered texture features in inland waterways exacerbate the difficulty of stereo feature matching. The second is the development of a Left-Right Consistency Refinement (LRR) module, which introduces the disparity from the target perspective in the middle stage of the network, using the pixel-level difference between the left and right disparities as a soft attention mechanism to alleviate the problem of error-prone areas being magnified during the upsampling process. In addition, We enhance LTNet, with knowledge distillation, leveraging CroCo-Stereo\cite{weinzaepfel2023croco} as our teacher network to achieve the optimized final model.

To demonstrate the superiority of our method, Fig. \ref{fig:comparison} shows the comparison between LTNet and state-of-the-art methods on the USVInland and the latest large-scale Spring dataset. It can be observed that LTNet achieves excellent performance on both datasets and strikes a delicate balance between deployment cost and prediction accuracy. Fig. \ref{fig:describe} illustrates examples of stereo matching in inland waterways using our method and the corresponding results obtained through the method of knowledge distillation. Since our research does not focus on the depth of the water surface, in the demonstration images, the water areas are removed using red and navy blue. The direct use of LTNet for stereo matching results in unclear object edges and prediction errors in distant scenes, which may be due to the sparse ground truth annotations and limited samples in the USVInland dataset. After the introduction of knowledge distillation, more accurate results can be obtained, thereby effectively addresses the overfitting risks brought about by the scarcity of training data.

In summary, our main contributions are listed below:

\begin{itemize}
    \item A novel geometric target volume is designed to encode the geometric information within the target features, in order to achieve higher matching accuracy.
    
    \item The Left-Right Consistency Refinement module is proposed to reduce mid-network mispredictions through similarity attention and multi-scale convolution.
    
    \item Knowledge distillation is first applied to stereo matching in inland waterways, and the resulting lightweight model with only 3.7M parameters achieves excellent performance on the USVInland dataset.
    
    \item Our method achieves superior results on the latest large-scale Spring benchmark, demonstrating its strong applicability across a variety of scenarios.
\end{itemize}

The rest of the paper is organized and detailed below. Section \ref{sec:related} details the related research work in learning-based stereo matching and knowledge distillation, Section \ref{sec:method} describes the workflow for each step of our LTNet method. Section \ref{sec:experiments} gives a detailed overview of the experiment. In the last section, we summarize and look forward to our work.

\section{Related Work}
\label{sec:related}

\subsection{Learning-Based Stereo Matching}

Compared with methods using hand-crafted features \cite{hirschmuller2007stereo, zhang2009cross, sun2011stereo}, end-to-end learning-based stereo matching methods have demonstrated superior performance. However, they also face challenges in deploying on resource-constrained edge devices due to the increasing scale of the networks. The precise and small-scale approach focuses on constructing lightweight 4D cost volumes (dimensions include channel, disparity, height, and width) and performing simple disparity processing, thus achieving a delicate balance between deployment cost and predictive accuracy.

To construct lightweight 4D cost volumes, some methods utilize techniques such as correlation \cite{bangunharcana2021correlate} and the concatenation of low-resolution features \cite{xu2021bilateral, dai2021adaptive}. Recently, more efforts have been concentrated on improving the accuracy of the information contained in limited-scale 4D cost volumes by mixing various factors. On the one hand, the combination cost volume approach is utilized. For instance, CFNet \cite{shen2021cfnet} improves the network's receptive field by fusing cost volumes from multiple resolutions of GwcNet \cite{guo2019group} cost volume. Afterwards, DCVSMNet \cite{tahmasebi2024dcvsmnet} attempts to use both group-wise correlation volume and norm correlation cost volume to form smaller volume in place of a single large cost volume. On the other hand, some advancements innovatively employ correlation as an attention mechanism to filter cost volumes, thereby generating efficient cost volumes. ACVNet \cite{xu2022attention} adopts this strategy, enhancing concatenation volumes by using multi-scale correlation volumes obtained through group-wise as cost attentions. A simpler approach, CGI-Stereo \cite{xu2023cgi} achieves lightweight cost volume construction by directly using correlation volumes as attention weights to enhance the matching of key areas in the reference features.

To achieve an efficient stereo matching network, in addition to constructing a lightweight 4D cost volume, it is also necessary to perform simple disparity processing on the predicted disparities. Simple disparity processing techniques help to enhance the performance of network without incurring a substantial computational load, one effective method of which is the left-right consistency check. This involves imposing uniqueness constraints on disparities by referencing information from the opposite perspectives. This is done to effectively eliminate unreliable areas in disparity estimation by creating confidence maps that reflect the results of these checks. In unsupervised model \cite{zhou2017unsupervised}, the predicted right disparity is used to warp to the left disparity using a forward warp function. The differences between the warped left disparity and the predicted left disparity are then thresholded to produce a disparity confidence map that guides the network training. LRCR \cite{jie2018left} generates left and right disparities cyclically using ConvLSTM and compares them to form an error map. And a soft attention mechanism is incorporated to selectively refine unreliable areas more effectively by utilizing the error map. Furthermore, PASMnet \cite{wang2020parallax} constructs cascading modules based on parallax-attention mechanism along the epipolar line and uses left-right consistency and cycle consistency checks for disparity results.

\subsection{Knowledge Distillation} 

In recent years, large-scale deep neural networks with high accuracy and generalization performance have achieved significant success in various fields. However, the massive number of parameters and computational requirements pose significant challenges for their deployment and real-time inference on resource-constrained devices \cite{gou2021knowledge}. To address this issue, the concept of model compression was first introduced under the research \cite{bucilua2006model} and later proposed the knowledge distillation method \cite{hinton2015distilling}. Knowledge distillation typically employs a student-teacher learning framework, where the student model is trained not only on the labels but also on the outputs of the teacher model. The aim is for the student model to imitate the teacher in order to achieve similar or even superior performance. From a training perspective, knowledge distillation can be categorized into three types: offline distillation \cite{hinton2015distilling}, online distillation \cite{anil2018large, zhang2021student}, and self-distillation \cite{hou2019learning, zhang2019your}.

In the field of depth estimation, high-accuracy stereo networks commonly serve as teachers to guide the training of small-scale depth estimation networks. The study \cite{guo2018learning} proposed using stereo matching network to learn from synthetic data and generate proxy disparity labels to supervise the monocular network. The monoResMatch \cite{tosi2019learning} proposed using traditional stereo algorithms, such as Semi-Global Matching, to obtain proxy ground truth annotation for training more accurate monocular networks. Furthermore, we recognize that recent stereo matching networks have achieved robust accuracy and generalization capabilities due to over parameterization. Although this leads to large models that are nearly impossible to deploy on embedded platforms, it makes it feasible to obtain lightweight and accurate small stereo matching networks through knowledge distillation.

\section{Method}
\label{sec:method}

The complex inland river environment and the long-distance space can cause texture blurring and small parallax in the stereo pair, which is composed of reference images and target images. To address these challenges, we introduce the concept of efficient matching driven by the target image. Therefore, a target-driven lightweight method suitable for USVs in inland waterways is proposed, named LTNet. It is worth noting that since we predict the left disparity of the stereo pair, the term "target" in this paper refers to the right-hand perspective. In this section, we first briefly introduce the architectural design of the proposed LTNet, followed by a detailed explanation of its principal components.

\subsection{Overview of LTNet}

\begin{figure*}[tbp]
\centering
\subfloat{\includegraphics[width=.9\linewidth]{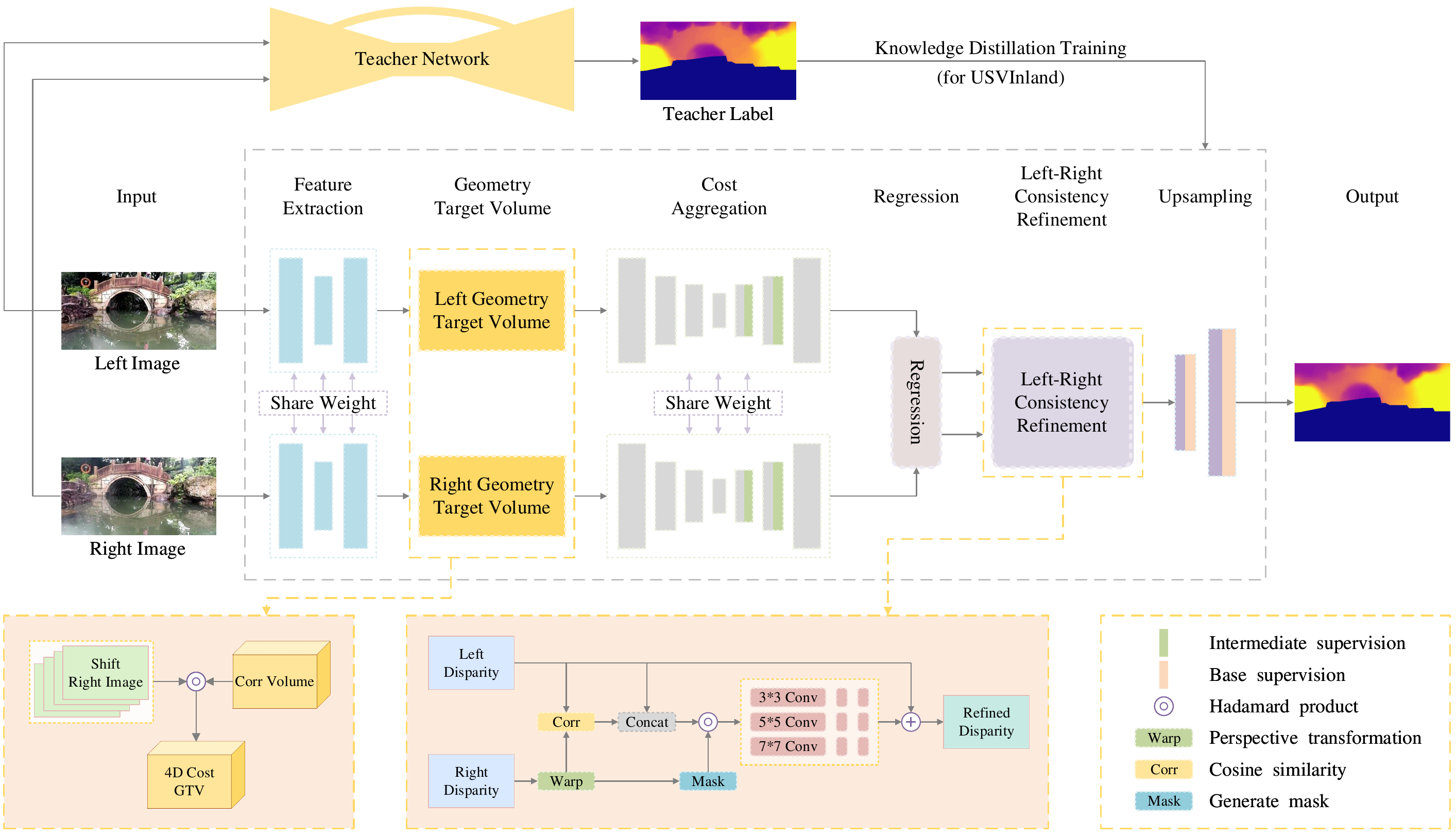}}
\caption{Architecture overview of the proposed LTNet. A stereo pair is fed into a feature extractor with shared weights, and 4D Geometry Target Volumes (GTVs) are constructed for the left and right branches respectively. These costs are fed into a 3D CNN network with shared weights and intermediate supervision to obtain the aggregated costs for both sides. After regressing to disparity maps respectively, the Left-Right Consistency Refinement (LRR) module is used to perform left-right consistency check and refinement. Finally, the disparity is upsampled to produce the predicted results. For the USVInland dataset, we further enhance the matching performance of lightweight models through knowledge distillation training.}
\label{fig:overview}
\end{figure*}

The overall architecture of the proposed network is shown in Fig. \ref{fig:overview}. It primarily comprises six modules: feature extraction, Geometry Target Volume (GTV) construction, cost aggregation, regression, Left-Right Consistency Refinement (LRR), and upsampling. The details of each network component are described in our appendix.

Specifically, multi-scale features are first extracted from the left and right images using a shared weights approach. The feature extractor employs a pre-trained MobileNetV2 on ImageNet as the backbone. Then, GTVs as lightweight 4D cost volumes are constructed separately for the left and right branches utilizing the 1/4 resolution features (Section \ref{subsec:GTV}). Subsequently, shared cost aggregation is performed on two lightweight cost volumes using 3D convolution, and multi-scale context fusion is combined following \cite{xu2023cgi}. In this context, we employ intermediate supervision for the aggregated costs at 1/16 and 1/8 scales to constrain the learning process. For the regression, top-k soft argmin regression \cite{bangunharcana2021correlate} is performed at 1/4 resolution to obtain coarse disparities for the left and right views (Section \ref{subsec:regression}).

Afterward, we further process the low-resolution left coarse disparity. The LRR module is proposed to achieve disparity constraints, ultimately outputting disparity at 1/4 resolution (Section \ref{subsec:LRR}). For upsampling, we extract the 3x3 neighborhood of unary features from the left image, and then perform weighted nearest-neighbor interpolation to the original resolution following \cite{yang2020superpixel}. Subsequently, by combining the dilated refinement module following \cite{chabra2019stereodrnet}, the final disparity results are obtained. In this process, we supervise both the disparity at 1/4 resolution before upsampling and the final prediction.

\subsection{Geometry Target Volume} 
\label{subsec:GTV}

After utilizing the shared feature extractor to acquire the features, a cost volume is constructed to support the subsequent matching computations. The traditional CNN methods using concatenation approaches \cite{kendall2017end, chang2018pyramid} and combination methods \cite{guo2019group, shen2021cfnet} can compensate for the insufficient modeling capabilities of the correlation layer methods \cite{mayer2016large}, but their redundant information representation and large channel dimensions will increase the operational burden of the network, which is unfavorable for deployment platforms with cost constraints such as USVs. Recent work has recognized that constructing a rich and concise cost volume can be achieved by filtering the feature volume using attention weights that include matching correlation \cite{xu2022attention, xu2023cgi}. In this context, the filtered reference image is rich in contextual information, while the target image is advantageous for fine textures due to its ability to assist in matching. However, the feature information of the target image is often weakened and even not directly utilized.

Additionally, for different scenarios, the design of the cost volume is often considered differently. In popular scenarios such as urban and indoor environments, the introduction of contextual features can significantly increase predictive advantages due to the simple geometry of man-made objects. On the other hand, in natural scenes with complex surfaces, such as forest environments, the cost depends on the matching information that is concatenated between pairs of images \cite{liu2022lanet}. However, in complex environments like inland river scenes, which include not only man-made objects such as bridges and riverbanks but also natural objects like plants on the shore and rocks, constructing a matching cost requires careful consideration of the potential relationships between stereo images. In order to construct accurate matching information for inland waterways within a limited volume, we propose the construction of a lightweight 4D Geometry Target Volume (GTV) rich in target information for the left and right network branches, serving as the efficient cost volume for our LTNet.

Stereo matching is the process of estimating the distance required to move a reference feature to a target location, where the target feature has the potential to reconstruct the geometry of the reference image. Therefore, we propose constructing a lightweight 4D cost volume that implicitly contains geometric information using disparity-level target features as the focus of attention. First, we construct a correlation volume \(V_{corr}\) using left and right feature maps at 1/4 resolution as,
\begin{equation}
\label{eqn:GTV_corr}
V_{corr}(d, x, y)=\frac{1}{N_{c}}\cdot\frac{<f_{l}(x, y),f_{r}(x-d, y)>}{\left\|f_{l}(x, y)\right\|_{2}\cdot\left\|f_{r}(x-d, y)\right\|_{2}},
\end{equation}
where \(d\) is disparity index, \(N_{c}\) is the number of channels, \(f_{l}\) and \(f_{r}\) represent the left and right feature maps, and \((x, y)\) denotes the pixel coordinate indices. Additionally, \(<\cdot,\cdot>\) denotes the inner product of two feature vectors, and \(\left\|\cdot\right\|_{2}\) represents the computation of the L2 norm. Afterward, we expand the channels of \(V_{corr}\) from 1 to 8 to form correlation attention weight \(A_{corr}\). By introducing the aforementioned similarity calculation, it becomes easier to aggregate the cost.

Then, we construct the target feature volume using the feature maps on the right side. Since the left disparity is predicted, the right features is identified as the target features. Specifically, the right feature maps at a resolution of 1/4 are compressed into 8 channels, represented as \(F_{r}\), and are shifted at each disparity levels. Finally, the target feature volume is filtered using the correlation attention weights \(A_{corr}\) to enhance the regions of correlation, resulting in the Geometry Target Volume \(V_{GT}\in\mathbb{R}^{B \times 8 \times D/4 \times H/4 \times W/4}\) as,
\begin{equation}
\label{eqn:GTV}
V_{GT}(\cdot,d,x,y)=A_{corr}\odot F_{r}(x-d, y),
\end{equation}
where \(\odot\) denotes the Hadamard product. The accordingly size of \(V_{GT}\) is \(N_{c}\times D/4 \times H/4 \times W/4\), D denotes the maximum of disparity, and the channel \(N_{c}\) is set to 8. Finally, we perform a simple aggregation on the obtained GTV with a convolutional kernel size of \(1 \times 5 \times 5\). By incorporating similarity measurements into the cost volume, neural network models are able to more accurately simulate the physical matching process, thereby enhancing their capability to model image matching locations.

The GTV encodes matching information and target features. By selectively focusing on the shifted target features, the geometric information of the target area can be used to help construct contextual semantics and reconstruct the structural information of the reference image. At the same time, due to the shift of the target image, the geometric structure of the reference image is incorporated into this target volume. We have enhanced the embedded structural information using correlation attention, thereby achieving the purpose of encoding costs. Ultimately, this method that employs target features to assist in reconstructing the geometric structure of the reference image aids in the construction of efficient cost volume.

\subsection{Left-Right Consistency Refinement}
\label{subsec:LRR}

Due to the presence of numerous complex natural textures in inland waterways, the simple monocular disparity matching process may struggle to handle detailed geometric information, leading to low-resolution predictions obtained during the mid-stage of network operation being easily mistaken due to texture blur and object occlusion, which can significantly adversely affect the subsequent up-sampling process of the network. To address these predictive errors, it is necessary to implement full-range constraints and corrective measures during the low-resolution phase of disparity estimation. In response to this, we propose the Left-Right Consistency Refinement (LRR) module with target disparity. This module employs an automated method for inspection to locate erroneous positions and performs focused refinement corrections.

Specifically, the warping operation is first used to achieve the perspective transformation that warps the right disparity to the left view. Then, shared point convolution is used to expand the left and right disparities into 6 channels, and consistency feature maps \( C_{lr} \) are obtained by calculating the similarity for each pixel across the channel dimension as,
\begin{equation}
\label{eqn:LRR_consis}
C_{lr}=\frac{<f^{1\times1}(d_{l}),f^{1\times1}(W(d_{r}))>}{\left\|f^{1\times1}(d_{l})\right\|_{2}\cdot\left\|f^{1\times1}(W(d_{r}))\right\|_{2}},
\end{equation}
where \(d_{l}\) and \(d_{r}\) represent the disparities of the left and right views, \(W(\cdot)\) is the perspective transformation, and \(f^{1\times1}\) is a \(1\times1\) convolution. The warping operation on the target disparity is achieved through scatter operator, which results in irregular hole regions in the outcome, manifesting as the absence of pixels in the generated reference disparity. However, due to the uniqueness constraint of disparity, the lack disparity corresponds exactly to the occluded areas. Subsequently, the occluded areas are extracted as a binary mask map. After connecting the left disparity and the consistency map \(C_{lr}\) on the channel dimension and expanding it to 8 channels, it is element-wise multiplied with the mask and then subjected to a convolution and a tanh activation to obtain the consistency attention \(A_{lr}\) as,
\begin{equation}\
\label{eqn:LRR}
A_{lr}=\tanh(f^{5\times5}(Concat\left\{d_{l}, C_{lr}\right\}\odot mask),
\end{equation}
where \(f^{5\times5}\) represents a \(5\times5\) convolution operation, which is used to aggregate disparity and similarity information locally, \(Concat\left\{\cdot,\cdot\right\}\) represents concatenation along the channel dimension, and the \(mask\) is a binary mask from \(W(d_{r})\). After obtaining consistency attention \(A_{lr}\), we use a residual refinement method to calibrate the disparity. First, we filter the left disparity \(d_{l}\) using \(A_{lr}\) to obtain the residual to be processed and expand the channel count to 8. Then, we process the residual using three layers of multi-scale convolution. For each layer, 2D convolutions with kernel sizes of 3, 5, and 7 are applied to the input to obtain residual maps with different receptive fields. After concatenating these on the channel dimension, a \(5\times5\) convolution for aggregation is utilized to achieve a more refined residual result with 8 channels, which is then input into the next multi-scale convolution layer. Ultimately, we compress the residual refinement results to one channel, add it to the left disparity \(d_{l}\), and apply a linear rectification function to obtain the refined disparity at 1/4 resolution.

Additionally, to aid in obtaining more accurate correlation map, we employ intermediate supervision in the cost aggregation process of both the left and right branches. This technique not only improves the network's performance but also provides more detailed left and right disparities, which are crucial for our automated inspection and refinement operations.

In summary, the proposed LRR module performs inspections and refinements to correct potentially erroneous estimates. It uses the left-right consistency of disparity to identify the location of errors. This involves imposing uniqueness constraints on disparities by referencing information from opposite perspectives, thereby creating attention maps. The areas described by the attention map allow our multi-scale convolution to effectively correct predictions and further refine the disparity. This flexible approach also benefits the disparity upsampling process, ultimately leading to extremely fine results.

\subsection{Disparity Regression}
\label{subsec:regression}

For the aggregated cost volume, we employ the top-k soft argmin regression strategy \cite{bangunharcana2021correlate} to obtain the disparity map. Firstly, the aggregated costs at each pixel are sorted, and the top k costs along with their corresponding disparity positions are selected. Subsequently, the soft argmin regression \cite{kendall2017end} is utilized to obtain the disparity map \(\hat{d}\) as,
\begin{equation}
\label{eqn:regression}
\hat{d}=\sum_{d=D_{\min}^{k}}^{D_{\max}^{k}}d\times\sigma\left(-c_{d}^{k}\right),
\end{equation}
where \(D_{\min}^{k}\) and \(D_{\max}^{k}\) represent the extreme values of the top-k candidate disparities \(D^{k}\), and \(c_{d}^{k}\) denotes the top-k candidate cost. In our LTNet, we set k=2. Additionally, the softmax operation \(\sigma\left(\cdot\right)\) is employed to normalize the cost along the disparity dimension into disparity probabilities.

\subsection{Loss Function}

Our loss function is divided into the basic loss \(\mathcal{L}_{B}\) and the intermediate loss \(\mathcal{L}_{I}\). The basic disparity loss supervises the final disparity at full resolution and the disparity refined through LRR at 1/4 resolution, defined as,
\begin{equation}
\label{eqn:loss_base}
\mathcal{L}_{B}=\lambda_{0}smooth_{L_{1}}\left(d_{0},d^{gt}\right)+\lambda_{1}smooth_{L_{1}}\left(d_{1},d^{gt}\right),
\end{equation}
where \(d^{gt}\) is ground truth disparity, \(d_{0}\) and \(d_{0}\) respectively denote the final disparity and the 1/4 resolution disparity. The coefficients are set as \(\lambda_{0} = 1.0\), \(\lambda_{1} = 0.3\). \(smooth_{L_{1}}\left(\cdot\right)\) calculates each disparity, which is widely applied in stereo matching, given as,
\begin{equation}
\label{eqn:loss_sml1}
smooth_{L_{1}}=\left\{
\begin{array}{ll}
0.5x^{2}, & \text{if}|x|<1\\
|x|-0.5, & \text{otherwise}
\end{array}.\right.
\end{equation}

To constrain the learning process and enhance the network's shallow disparity learning ability, we use intermediate supervision separately on the left and right cost aggregation processes. This supervision targets the costs generated at 1/16 and 1/8 resolutions. Here, we employ transpose convolution to upsample both costs to 1/4 resolution before regressing to obtain the supervised disparities. The intermediate supervision loss is defined as,
\begin{equation}
\label{eqn:loss_intermediate}
\begin{split}
\mathcal{L}_{I}=\sum_{i=1}^{2}\left[\lambda_{2}smooth_{L_{1}}\left(d_{i}^{8},d^{gt}\right)+\lambda_{3}smooth_{L_{1}}\left(d_{i}^{16},d^{gt}\right)\right],
\end{split}
\end{equation}
where \(i\) denotes the choice of either the left or right aggregation branch for supervision, \(d^{8}\) and \(d^{16}\) respectively denote the regressed disparities from costs at 1/8 and 1/16 resolutions.  The coefficients for intermediate supervision are set as \(\lambda_{2} = 0.2\), \(\lambda_{3} = 0.1\). Additionally, the \(d^{gt}\) used when supervising the right branch is obtained by warping the left true disparity. Ultimately, the sum of the basic loss \(\mathcal{L}_{B}\) and the intermediate loss \(\mathcal{L}_{I}\) constitutes the loss for LTNet.

\subsection{Knowledge Distillation Training}

In our experiments, it was observed that training stereo networks with the USVInland dataset typically underperforms in various lightweight algorithms, which may be due to the insufficiency of sample quantities. Additionally, compared to the commonly used road vehicle datasets \cite{geiger2013vision, menze2015object}, the ground truth in USVInland is more sparse. To alleviate the overfitting problem about by the scarcity of training data, we propose a two-stage offline distillation method for training an end-to-end stereo matching model with a straightforward pipeline. It should be noted that our knowledge distillation training is conducted solely on the USVInland dataset.

Here we define CroCo-Stereo \cite{weinzaepfel2023croco} as the teacher while LTNet as the student. Due to CroCo-Stereo employs a network architecture that integrates the Vision Transformer \cite{dosovitskiy2020image} with a dense prediction transformer \cite{ranftl2021vision} head. Unlike other stereo networks, this architecture does not rely on pre-set disparity sequence regression results. Therefore, we directly utilize the output disparity as the teacher knowledge. And according to our experiments, learning the proxy labels of teacher and ground truth simultaneously is not a correct training strategy. Thus, we ensure that the student prioritizes learning from teacher through a two-stage method, while also being able to correct inaccurate knowledge.

For our two-stage approach, during the first stage, we learn hard labels derived from the regression of the teacher model. Here, since the disparity map output by the stereo teacher network is dense, the student can learn the areas that are not labeled with ground truth under the guidance of the teacher. For the second stage, ground truth is used to fine-tune the model again. Thanks to the exemplary guidance provided by the teacher, our model converge fast and further improve performance. In practical training, we employ a 5-fold cross-validation approach, obtaining 180 pairs of densely inferred disparity maps from CroCo-Stereo, which serve as our teacher labels after excluding the water surface disparities.

\section{Experiments}
\label{sec:experiments}

In this section, we evaluate our method on the USVInland dataset \cite{cheng2021we}. Due to the absence of prior work on stereo matching involving this dataset, we reproduce advanced algorithms as baselines. Furthermore, the latest large-scale Spring dataset \cite{mehl2023spring} is employed to evaluate the performance of our method across a variety of scenes. Ultimately, the superior performance on both the USVInland and Spring datasets demonstrates the advancement of our method in inland waterways and the applicability in diverse scenes.

\subsection{Datasets and Evaluation Metrics}

\begin{itemize}[leftmargin=0pt, itemindent=1.5em]
    \item Scene Flow: A large synthetic dataset is used to pre-train our LTNet at a resolution of \(960 \times 540\), including 35,454 pairs of training images and 4,370 pairs of testing images, using the Finalpass setting. The end-point error (EPE) is utilized as the evaluation metric, representing the average pixel discrepancy between the true values and the predicted disparity.

    \item USVInland: A real-world dataset collected by an unmanned surface vehicle is used to fine-tune our LTNet, which uses LiDAR to capture sparse ground truth, includes 180 pairs of inland waterway surface stereo images with a resolution of \(640 \times 320\). The evaluation metrics are based on the percentage of errors at disparity thresholds ranging from 1 to 3, as well as the end-point error.

    \item Spring: A high-detail dataset based on rendered scenes from the open-source movie "Spring", with a resolution of \(1920 \times 1080\), including 5,000 pairs of training images and 1,000 pairs of testing images. The evaluation metrics include the 1 pixel outlier rate, the percentage of outlier D1, and the absolute error. For D1, an error of less than 3 pixels or 5\% of the true disparity is considered as correctly estimated.
\end{itemize}

\subsection{Implementation Details}

We implement our network with PyTorch and employ Adam \cite{kingma2014adam} optimizer(\(\beta_{1} = 0.9\), \(\beta_{2} = 0.999\)). The batch size is set to 8 for training on a single NVIDIA RTX 4090 GPU.

On the Scene Flow dataset, we crop the stereo images to \(H = 256\) and \(W = 512\). We first train from scratch for 32 epochs, then fine-tune for an additional 32 epochs. The initial learning rate is set to 0.001 and is decayed by 2 times after 10, 14, 18, 22, and 26 epochs.

On the USVInland dataset, we randomly divide all stereo pairs into training set and validation set in an 8:2 ratio, fine-tune the pre-trained Scene Flow model for 600 epochs. The initial learning rate is set to 0.001, and decayed it by tenfold after 300 epochs. For the distillation training, we first train for 300 epochs with a learning rate of 0.001 using teacher labels on the pre-trained Scene Flow model. Afterwards, we fine-tune for another 50 epochs on the best-performing model using ground truth, with a learning rate of 0.0001. For performance experiments, 5-fold cross-validation is implemented to comprehensively evaluate the model performance, which is not used in ablation experiments.

\subsection{Baselines}

\begin{table*}[tbp]
\def \alambicl {\includegraphics[width=0.013\linewidth]{figs/alembic-crop.pdf}\xspace}
\centering
\caption{Results of the 5-fold cross-validation on the USVInland dataset and the EPE metrics on Scene Flow. The symbol $\alambicl$ refers to the model trained using distillation. \textbf{Bold}: Best, \underline{Underscore}: Second best.}
\begin{tabular*}{1.\linewidth}{@{\extracolsep{\fill}}clccccccccc@{}}
\toprule
                             & \multirow{2}{*}{Method}                             & \multirow{2}{*}{Venue} & \multirow{2}{*}{Params} & \multirow{2}{*}{MACs} & \multirow{2}{*}{Runtime} & \multicolumn{4}{c}{USVInland}                                              & Scene Flow       \\ \cline{7-10} \cline{11-11}
                             &                                                     &                        &                         &                       &                          & error1            & error2           & error3           & EPE              & EPE              \\ \midrule
\multirow{2}{*}{Early work}  & SGM \cite{hirschmuller2007stereo}                   & TPAMI2007              & \_                      & \_                    & \_                       & 48.80             & 19.14            & 12.57            & \_               & \_               \\
                             & PSMNet \cite{chang2018pyramid}                      & CVPR2018               & 5.2M                    & 194.5G                & 63.1ms                   & 19.10             & 7.38             & 4.33             & 0.78             & 1.09             \\ \midrule
\multirow{3}{*}{Large-scale} & IGEV-Stereo \cite{xu2023iterative}                  & CVPR2023               & 12.6M                   & 505.6G                & 137.7ms                  & 14.67             & 6.47             & 3.88             & 0.63             & 0.47             \\
                             & CroCo-Stereo \cite{weinzaepfel2023croco}            & ICCV2023               & 437.4M                  & 663.6G                & 45.8ms                   & 14.73             & 6.03             & 3.56             & 0.66             & \_               \\
                             & CGF-ACV \cite{xu2023cgi}                            & arXiv2023              & 10.6M                   & 240.9G                & 101.7ms                  & 16.14             & 6.12             & 3.47             & 0.69             & 0.52             \\ \midrule
\multirow{4}{*}{Lightweight} & MSNet3D \cite{shamsafar2022mobilestereonet}         & WACV2022               & 1.8M                    & 88.0G                 & 24.2ms                   & 18.16             & 6.60             & 3.58             & 0.73             & 0.80             \\
                             & CoEx \cite{bangunharcana2021correlate}              & IROS2021               & 2.7M                    & 18.6G                 & 7.3ms                    & 18.16             & 6.65             & 3.58             & 0.74             & 0.69             \\
                             & CGI-Stereo \cite{xu2023cgi}                         & arXiv2023              & 3.5M                    & 17.6G                 & 11.1ms                   & 17.35             & 6.25             & 3.55             & 0.73             & \underline{0.64} \\
                             & LTNet(ours)                                         &                        & 3.7M                    & 45.0G                 & 14.7ms                   & \underline{15.66} & \underline{6.13} & 3.50             & \underline{0.69} & \textbf{0.60}    \\ \midrule
\multirow{4}{*}{Distilled}   & MSNet3D\alambic \cite{shamsafar2022mobilestereonet} & WACV2022               & 1.8M                    & 88.0G                 & 24.2ms                   & 17.30             & 6.48             & 3.68             & 0.71             & 0.80             \\
                             & CoEx\alambic \cite{bangunharcana2021correlate}      & IROS2021               & 2.7M                    & 18.6G                 & 7.3ms                    & 17.33             & 6.37             & \underline{3.45} & 0.72             & 0.69             \\
                             & CGI-Stereo\alambic \cite{xu2023cgi}                 & arXiv2023              & 3.5M                    & 17.6G                 & 11.1ms                   & 17.09             & 6.48             & 3.51             & 0.74             & \underline{0.64} \\
                             & LTNet\alambic(ours)                                 &                        & 3.7M                    & 45.0G                 & 14.7ms                   & \textbf{15.36}    & \textbf{6.01}    & \textbf{3.44}    & \textbf{0.67}    & \textbf{0.60}    \\ \bottomrule
\end{tabular*}
\label{tab:quantity}
\end{table*}

We compare our proposed LTNet with two early works, three large-scale methods, and three lightweight methods, briefly introduced as follows.

\begin{itemize}[leftmargin=0pt, itemindent=2em]
    \item SGM \cite{hirschmuller2007stereo}: An early work employs semi-global path-wise optimization and pixel-wise mutual information based matching cost to complete stereo matching.

    \item PSMNet \cite{chang2018pyramid}: An early work employs a Spatial Pyramid Pooling module and multiple stacked hourglass networks, combined with intermediate supervision to regularize the cost.

    \item IGEV-Stereo \cite{xu2023iterative}: A large-scale method constructs a Combined Geometry Encoding Volume to encode geometry and context information, as well as local matching details, and updates the disparity map through iterative indexing.

    \item CroCo-Stereo \cite{weinzaepfel2023croco}: A large-scale method employs pre-training of the Vision Transformer model using cross-view completion as a pretext task, and incorporates the Dense Prediction Transformer head to output disparity maps.

    \item CGF-ACV \cite{xu2023cgi}: A large-scale method incorporates Context and Geometry Fusion into ACVNet, which utilizes multi-level adaptive weights to generate the 4D cost volume.

    \item MSNet3D \cite{shamsafar2022mobilestereonet}: A lightweight method employs efficient 3D convolution blocks from the MobileNet, which are introduced into the stereo network to aggregate the 4D cost volume.

    \item CoEx \cite{bangunharcana2021correlate}: A lightweight method constructs a real-time stereo matching network using Guided Cost Volume Excitation and a top-k soft-argmin disparity regression method.

    \item CGI-Stereo \cite{xu2023cgi}: A lightweight method constructs an efficient network using Context and Geometry Fusion blocks to adaptively perform information fusion in cost aggregation.
\end{itemize}

\subsection{Performance on USVInland Dataset}

We pre-train models using the Scene Flow dataset and fine-tune them on the USVInland dataset to evaluate the performance of our LTNet against advanced methods. Due to the relatively small amount of USVInland data, we employ a 5-fold cross-validation method to obtain the final results. Specifically, we randomly divide the USVInland stereo dataset into five parts, each serving as a test subset. For each subset, the training set comprises the remaining four parts to maintain an 8:2 training to validation set ratio. This results in five subsets used for cross-validation, and we use the average of metrics to obtain the final quantitative results.

Considering the increasing deployment of USVs for autonomous tasks in waterways, these aquatic devices often face cost constraints. Simultaneously, due to the real-time safety and sensing requirements of the operational equipment, stereo matching is also a task that is extremely sensitive to time. Therefore, the design of lightweight stereo matching networks faces dual challenges of computational cost and runtime. In this work, we design a target-driven high-precision lightweight algorithm called LTNet for USVs platforms on inland waterways. LTNet uses a constrained parameter strategy with only 3.7M parameters, achieving the optimal balance between model size and performance. We present the error rates at different thresholds and the EPE metric obtained by our model and other advanced methods on the USVInland dataset in TABLE \ref{tab:quantity}, as well as the EPE metric on the Scene Flow dataset. The data for the SGM method comes from the paper \cite{cheng2021we}. The underscore notation in the CroCo-Stereo method is due to the inclusion of Scene Flow test samples in the official pre-training. And it should be noted that we mark the lightweight and the distilled models for comparison.

In the evaluation of lightweight structures, we not only assess the indirect metrics of the models but also evaluate direct metrics on the same computing platform to meet practical deployment requirements. Among these, the number of parameters and the MAC (multiply-accumulate) metric are used as indirect metrics, while the running time is used as a direct metric. Since the additional right branch only regresses low-resolution disparity and uses a highly parallel 3D convolutional aggregation method, the proposed LTNet, although it increases the computational load within an acceptable range, maintains an equal level of running time with other lightweight networks, which is sufficient to meet the cost and real-time requirements of USVs. It can be observed that, among the lightweight methods, our method achieved the best scores across all metrics. Furthermore, our LTNet demonstrates considerable competitiveness on the USVInland dataset when compared with large-scale models.

\begin{figure*}[tbp]
\centering
\subfloat{\includegraphics[width=\linewidth]{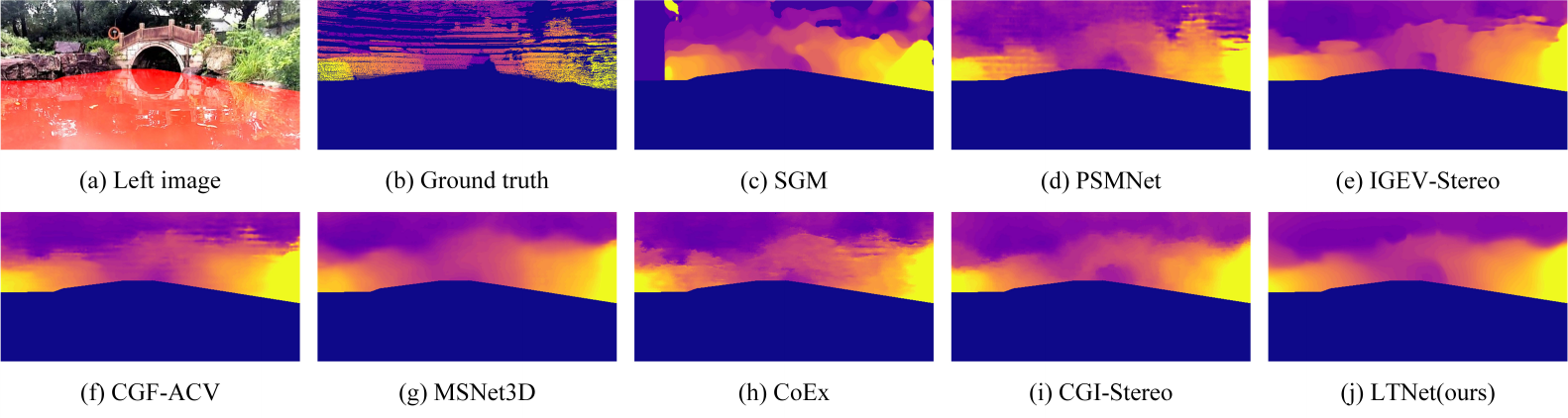}}
\caption{Results of disparity estimation for USVInland test image. Note that in our demonstrations, water areas have been removed using red and navy blue.}
\label{fig:quality}
\end{figure*}

Additionally, we qualitatively compare our algorithm with advanced methods on the USVInland dataset. In Fig. \ref{fig:quality}, we show the pseudo-color disparity results of USVInland test images obtained through inference by various algorithms. Please note that our research does not focus on the depth of the water surface. In our demonstration images, water areas have been removed using red and navy blue. As demonstrated in our examples, various algorithms are prone to matching errors in complex inland waterway scenes, which can severely interferes with the environmental perception of USVs. For instance, the results of the SGM algorithm have failure areas, the disparity representation of PSMNet is not sufficiently smooth, and the results from CGF-ACV and CoEx are prone to causing ripple-like error areas. In comparison, our LTNet produced more accurate results, performing better in detail and smoother disparity maps.

Lightweight methods fail to produce accurate results, partly due to their cost volume and the lack of constraint refinement. In terms of cost volume, lightweight methods typically only consider similarity information, as seen in MSNet3D and CoEx, which respectively acquire cost volumes through group-wise correlation and cosine similarity calculations. Although CGI-Stereo uses attention filtering to generate cost volumes, it tends to over-focus on the contextual information of reference features, which can result in insufficient modeling capability for matching costs. On the other hand, lightweight methods lack automatic constraints on predicted disparities and further corrections. For CoEx and CGI-Stereo, the clever use of low-resolution regression followed by upsampling avoids the high computational costs of large volumes. However, errors in low-resolution predictions are directly magnified, which severely impacts the performance of network.

By contrast, based on our target-driven concept, the network utilize the geometric features of the target image and the disparity from the target perspective to assist in the realization of matching. Specifically, our LTNet reconstructs the geometric structure of the reference image by moving target image features, constructing an efficient cost volume, which is the cornerstone of our network's lightness and accuracy. Moreover, by introducing the left-right consistency of disparities to assist in the mid-term refinement of predictions, this approach automatically identifies the erroneous areas in the predicted map and subsequently focuses on correcting them. Based on the optimization of the cost volume and low-resolution disparity, LTNet achieves a delicate balance between model size and performance.

Furthermore, we have introduced a knowledge distillation training method to improve disparity prediction in inland waterway scenes and better serve USVs. By prioritizing learning from teacher knowledge derived from CroCo-Stereo and utilizing ground truth to correct inaccuracies, the student lightweight model can achieve performance improvements. As shown in TABLE \ref{tab:quantity}, if trained with knowledge distillation, LTNet performs even better and continues to significantly lead among lightweight methods. This demonstrates the effectiveness of our knowledge distillation method in inland waterways, as well as the potential of our accurate lightweight model in practical application scenes.

\subsection{Ablation Study}

\begin{figure*}[tbp]
\def \alambicm {\includegraphics[width=0.07\linewidth]{figs/alembic-crop.pdf}\xspace}
\centering
\subfloat{\includegraphics[width=\linewidth]{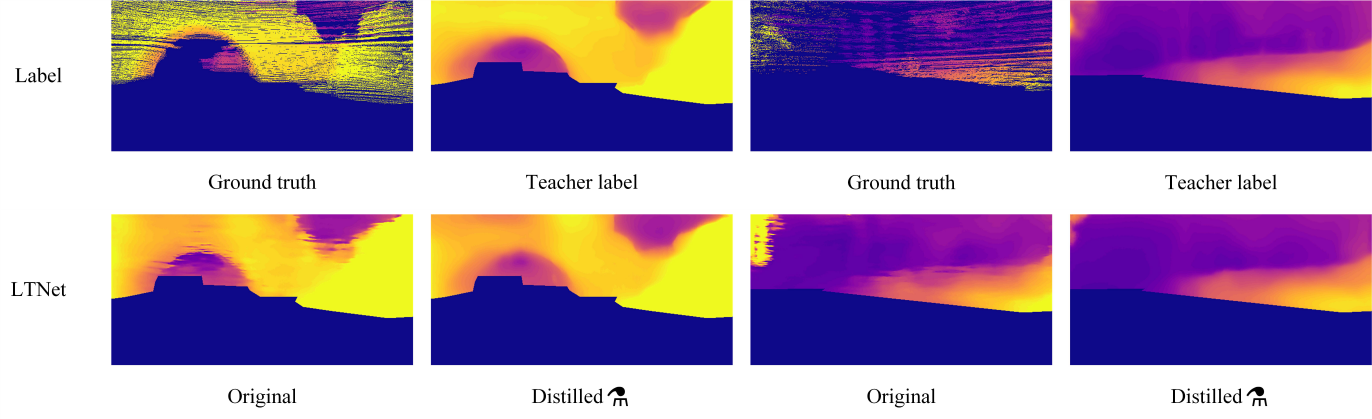}}
\caption{Comparison of knowledge distillation training results on USVInland test images. The symbol $\alambic$ refers to the model trained using distillation. Our teacher labels utilize the output disparity from CroCo-Stereo \cite{weinzaepfel2023croco}. Note that water areas in the labels and results have been removed using navy blue.}
\label{fig:distil}
\end{figure*}

In this section, we conduct ablation study to investigate the influence of the proposed modules. These experiments are performed using the USVInland dataset, with all models trained from scratch. As shown in TABLE \ref{tab:ablation_1}, We conduct replacement training of the cost volume for our model to demonstrate the effectiveness of our Geometry Target Volume (GTV). In another ablation study, we integrated Left-Right Consistency Refinement and intermediate supervision mechanism into the baseline model for constructing the GTV, as shown in TABLE \ref{tab:ablation_2}. Furthermore, to evaluate the effectiveness of our distillation training, we show the results using different training settings on the proposed LTNet, as shown in TABLE \ref{tab:quantity} and Fig. \ref{fig:distil}.

\begin{table}[tbp]
\centering
\caption{Effectiveness of GTV, comparing five cost constructions for ablation experiment. \textbf{Bold}: Best, \underline{Underscore}: Second best.}
\begin{tabular}{lccccccc}
\toprule
Method            & Left       & Right      & Corr       & error1            & error2           & error3           & EPE               \\ \midrule
Concat            & \checkmark & \checkmark &            & 39.40             & 17.32            & 8.59             & 1.173             \\
Corr              &            &            & \checkmark & \underline{21.49} & \underline{7.56} & \underline{3.41} & \underline{0.743} \\
Combine           & \checkmark & \checkmark & \checkmark & 23.61             & 8.49             & 3.78             & 0.791             \\
AFV               & \checkmark &            & \checkmark & 21.76             & 7.61             & 3.42             & \underline{0.743} \\
GTV(ours)         &            & \checkmark & \checkmark & \textbf{21.02}    & \textbf{7.38}    & \textbf{3.26}    & \textbf{0.733}    \\ \bottomrule
\end{tabular}
\label{tab:ablation_1}
\end{table}

\subsubsection{Geometry Target Volume}

In this study, we evaluated five different methods for constructing cost volumes, each incorporating various components. The first method involves using concatenation volumes obtained by channel level connection of extracted left and right features, which was first used by GC-Net \cite{kendall2017end}. Since this volume does not contain direct correlation information, it requires the model's aggregation network to consume resources to learn the stacked feature similarities, making it unsuitable for use in lightweight networks. The second construction method extends the correlation volume from DispNetC \cite{mayer2016large} by expanding the channels. This correlation operation involves calculating the cosine similarity between the left and right features. The third method constructs a combine volume based on both concatenation and correlation approaches by directly multiplying the two volumes. The penultimate method is the Attention Feature Volume (ATV) proposed by CGI-Stereo \cite{xu2023cgi}, which uses the correlation volume as an attention mechanism to filter the left features without directly utilizing the right features. The last method is our proposed Geometry Target Volume.

We compare our GTV based on target features with four other cost volumes containing different matching information. The experimental results in TABLE \ref{tab:ablation_1} show that for a lightweight matching network composed of a limited number of 3D convolutions, it is difficult to learn matching patterns from concatenation volume lacking correlation computation operations, resulting in metrics significantly higher than those of other models. When we add correlation to the concatenation volume, the resulting combine volume model reduces the EPE metric to 0.791 pixels, but ultimately performs poorly due to cost redundancy. Our novel approach enriches the cost volume with abundant target features and reconstructs it based on the geometric structure of the reference image. In models using the correlation volume, AFV, and GTV, constructing the cost volume with our method outperforms other implemented methods across all metrics, making it more suitable for lightweight stereo matching networks.

\subsubsection{Left-Right Consistency Refinement}

In this ablation study, we constructed a baseline model using the Geometry Target Volume based on LTNet and gradually added the Left-Right Consistency Refinement (LRR) module and intermediate supervision (IS) mechanism to demonstrate the effectiveness of our method, as shown in TABLE \ref{tab:ablation_2}. When we added LRR and intermediate supervision separately, the model's EPE metric respectively decreased by 0.013 and 0.017 pixels, resulting in only slight improvements. However, when both LRR and intermediate supervision were used together, EPE metric improved by 0.026 pixels, and almost all other metrics showed significant improvements as well.

\begin{table}[tbp]
\centering
\caption{Effectiveness of LRR, evaluation of LTNet under different settings. \textbf{Bold}: Best, \underline{Underscore}: Second best.}
\begin{tabular}{lcccccc}
\toprule
Method            & LRR        & IS         & error1            & error2           & error3           & EPE               \\ \midrule
Baseline          &            &            & 21.63             & 7.79             & 3.59             & 0.759             \\
LRR               & \checkmark &            & \textbf{21.02}    & 7.78             & 3.67             & 0.746             \\
IS                &            & \checkmark & \underline{21.21} & \underline{7.64} & \underline{3.38} & \underline{0.742} \\
LRR+IS(ours)      & \checkmark & \checkmark & \textbf{21.02}    & \textbf{7.38}    & \textbf{3.26}    & \textbf{0.733}    \\ \bottomrule
\end{tabular}
\label{tab:ablation_2}
\end{table}

This is because our LRR module performs consistency refinement based on left-right aggregation branches with shared weights. We calculate the similarity between the predictions of the left and right branches to identify the mid-term errors in the network and perform multi-scale refinement to add left-right consistency to the disparity. When identifying prediction errors, it is necessary to obtain accurate left and right disparities to optimize consistency attention accurately. By introducing an intermediate supervision mechanism to separately supervise the left and right aggregation processes, we can obtain more accurate low-resolution left-right disparity consistency attention in the middle of the network, guiding the network to focus on refining the target.

\begin{figure*}[tbp]
\centering
\subfloat{\includegraphics[width=\linewidth]{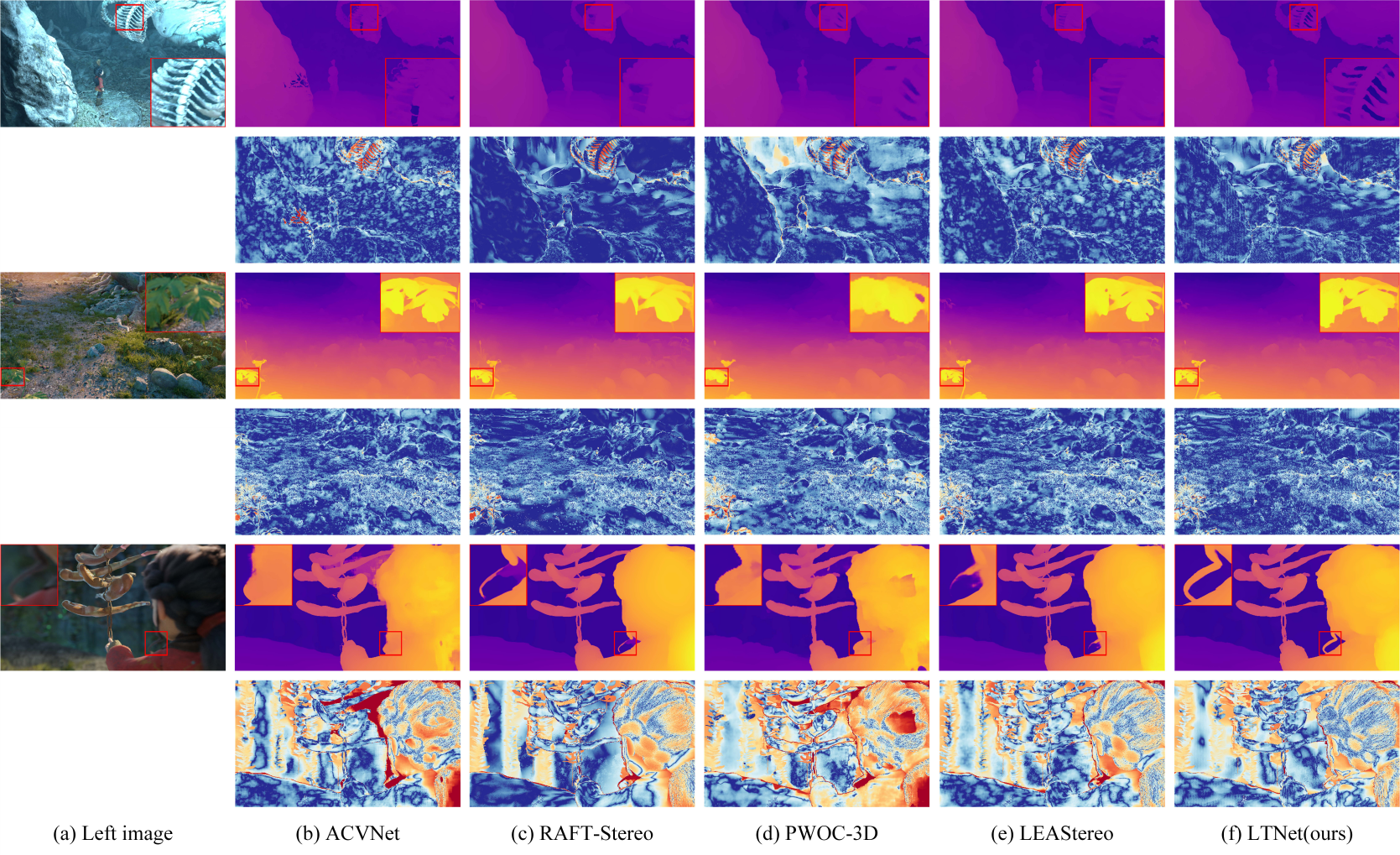}}
\caption{Results for Spring test images. Error in orange corresponds erroneous prediction. Details of the image are marked with red frames and enlarged.}
\label{fig:spring}
\end{figure*}

\subsubsection{Knowledge Distillation Training}

To evaluate the effectiveness of our distillation training, we compare the disparity results of LTNet under different training setups on the USVInland dataset. We show the error rates at different thresholds and the EPE metric in TABLE \ref{tab:quantity}, which illustrates the quantitative improvements of LTNet after distillation training. It can be observed that after distillation training, our LTNet achieves improvements across all evaluation metrics on the USVInland dataset. This is due to our method of reconstructing the geometric structure of the reference image by moving target features within the cost volume, allowing LTNet to comprehensively encode matching features between images. Such a structure enables our network to maintain its lightweight while possessing enhanced learning capabilities, better absorbing the rich knowledge output from the teacher network. Furthermore, knowledge distillation training leads to significant performance improvements in lightweight networks such as MSNet3D and CoEx, further corroborating the effectiveness of our distillation training approach.

In Fig. \ref{fig:distil}, we show the disparity results of different training settings on USVInland test images. It can be observed that in results without distillation training, ripple-like prediction errors frequently occur, which may attributed to the sparsity of ground truth in the USVInland dataset. After undergoing our distillation training, LTNet successfully removed these ripple errors and eliminated previous prediction errors at the edge of scene. Such improvements provide more reliable environmental information for deployed devices, which is crucial for enhancing the operational stability and safety of inland waterway USVs.

\subsection{Applicability Experiment}

To demonstrate that our method excels not only in inland waterway scenes but also in a broader range of scenarios, we conducted validations using the latest large-scale Spring dataset \cite{mehl2023spring}. It includes a variety of common natural scenes such as distant and close-up figures, bonfires, forests, and mountains. The dataset also permits the online submission of disparity field predictions for test images to impartially assess the performance of various methods. In addition, because the labels of the Spring dataset are sufficiently accurate, we do not perform knowledge distillation.

When training LTNet on the Spring dataset, images are cropped to a size of \(H = 256\), \(W = 512\). We utilize the pre-trained model on Scene Flow for two rounds of fine-tuning, each lasting 32 and 64 epochs respectively. The initial learning rate is set to 0.001 and is decayed by 2 times after 10, 14, 18, 22, and 26 epochs. After the training is completed, the results of LTNet are submitted online and compared with several published methods \cite{xu2022attention, lipson2021raft, saxena2019pwoc, cheng2020hierarchical}. As shown in TABLE \ref{tab:spring} and Fig. \ref{fig:spring}, the superior performance of LTNet confirms the broad adaptability of our approach across various scenes.

\begin{table}[tbp]
\centering
\caption{Results from Spring Benchmark Report. \textbf{Bold}: Best, \underline{Underscore}: Second best.}
\begin{tabular}{lccccc}
\toprule 
Method                                 & Venue       & Params & 1px               & D1               & Abs              \\ \midrule
ACVNet \cite{xu2022attention}          & CVPR2022    & 6.2M   & \textbf{14.77}    & \underline{5.35} & 1.52             \\
RAFT-Stereo \cite{lipson2021raft}      & 3DV2021     & 11.2M  & 15.27             & 8.63             & 3.03             \\
PWOC-3D \cite{saxena2019pwoc}          & IV2019      & 5.9M   & 18.23             & 5.89             & \underline{1.34} \\
LEAStereo \cite{cheng2020hierarchical} & NeurIPS2020 & 1.8M   & 19.89             & 9.19             & 3.88             \\
LTNet(ours)                            &             & 3.7M   & \underline{15.24} & \textbf{3.91}    & \textbf{0.83}    \\ \bottomrule
\end{tabular}
\label{tab:spring}
\end{table}

Among all compared methods, our LTNet ranks first in both the D1 and Abs metrics and second in the 1px metric. Due to our target-driven concept, the network extensively utilizes geometric features from the target image and optimizes disparity through the perspective of the target. This allows our method to clearly perform pixel matching in the details. In the demonstration, LTNet can distinctly discern the hollows of objects, fully restore the leaves on the hillside, and successfully match the face-framing hairs of the characters. This is attributed to our effective geometric reorganization and encoding design of target features, allowing the network to construct clear and efficient matching costs while maintaining high accuracy and lightweight characteristics. Additionally, thanks to ingenious Left-Right Consistency Refinement structure, the network is capable of autonomously identifying potential prediction errors and effectively executing corrections and optimizations, thereby significantly reducing the amplification of error positions in the disparity upsampling step. Furthermore, to facilitate use on resource-constrained platforms, our LTNet is designed with a restrained parameter scale. Compared to advanced algorithms such as ACVNet \cite{xu2022attention} and RAFT-Stereo \cite{lipson2021raft}, which have parameter counts 1.7 and 3.0 times larger than LTNet respectively, our approach is extremely lightweight and well-suited for application deployment in a variety of scenes.

\section{Conclusion}

To address the stereo matching problem for Unmanned Surface Vehicles (USVs) platforms in inland waterways, we proposed a target-driven lightweight stereo matching network called LTNet. This network achieved high-precision and smooth results effectively by reorganizing target features and utilizing target disparity for online refinement. Specifically, in the cost construction, we effectively encoded geometric information through the introduction of right-shift features and the correlation filtering strategy. Additionally, the Left-Right Consistency Refinement module was developed to identify and optimize intermediate erroneous predictions. In the context of inland waterways, knowledge distillation was introduced to further enhance the performance of lightweight models. We conducted quantitative and qualitative assessments on the USVInland dataset, and the results demonstrated that our LTNet was capable of accurately handling challenging stereo matching in waterway scenes. Furthermore, the competitive performance on the latest large-scale Spring benchmark demonstrated the multi-scenario applicability of our method. In our future work, we aim to extend our approach for stereoscopic video \cite{pan2020multi} and explore depth estimation tasks in other specific application scenarios, including underwater \cite{ye2019deep, ye2023underwater} and indoor \cite{wei2021iterative, li2022monoindoor++} environments.

\section*{Acknowledgments}
This work was supported by the National Natural Science Foundation of China under Grant 62106002.

\bibliography{references}
\bibliographystyle{IEEEtran}

\vspace{11pt}
\vfill

\end{document}